\documentclass[letterpaper, 10 pt, conference]{ieeeconf}
\IEEEoverridecommandlockouts

\usepackage{amsmath,amsfonts,amssymb,color,amsthm}
\usepackage{algorithm}
\usepackage{algpseudocode}
\usepackage{array}

\usepackage{graphicx}
\usepackage{cite}
\usepackage{footmisc}
\usepackage{mathtools}
\usepackage{bbm}
\usepackage{bm}
\usepackage{comment}
\usepackage{dsfont}

\usepackage{mathtools}
\usepackage{medmath}
\usepackage{tikz}
\usetikzlibrary{automata, positioning, arrows.meta}
\usepackage{relsize}
\usetikzlibrary{shapes,arrows}
\usepackage{cite}

\usepackage[colorlinks]{hyperref}
\definecolor{mygreen}{rgb}{0.0, 0.5, 0.0}
\definecolor{winered}{rgb}{0.7,0,0}
\definecolor{myblue}{rgb}{0,0,0.8}

\usepackage{scalerel}
\usepackage{xcolor}

\DeclarePairedDelimiterX{\norm}[1]{\lVert}{\rVert}{#1}

\newcommand{\mc}{\mathcal}

\AtBeginDocument{%
  \hypersetup{
    citecolor=myblue,
    linkcolor=winered}}

\newcommand{\smallsquare}{\scalebox{0.7}{$\blacksquare$}}

\DeclarePairedDelimiter\ceil{\lceil}{\rceil}
\DeclareMathOperator*{\argmin}{\arg\!\min}

\newtheorem{problem}{Problem}

\newtheorem{theorem}{Theorem}
\newtheorem{corollary}{Corollary}
\newtheorem{lemma}{Lemma}

\newtheorem{assumption}{Assumption}

\begin{document}
\title{\LARGE\bf Adaptive Memory for Online Reinforcement Learning\\ under Sparse Model Changes}
\author{Sreejeet Maity, Sukanya Modak, and Leonardo F. Toso %
\thanks{Sreejeet Maity is with the Department of Electrical and
Computer Engineering, North Carolina State University. Sukanya Modak is with the Department of Electronics and Communication Engineering, RCC Institute of Information Technology. Leonardo F. Toso is with the Department of Electrical Engineering, Columbia University.
Emails:~\texttt{\textcolor{winered}{smaity2@ncsu.edu}},
\texttt{\textcolor{winered}{ece2024065@rcciit.org.in}}, \texttt{\textcolor{winered}{leonardo.toso@columbia.edu}.}}}
\date{}
\maketitle
\thispagestyle{empty}
\pagestyle{empty}
\begin{abstract}
We study the discounted reinforcement learning (RL) problem in a setting where the underlying environment changes between deployments. Each deployment
provides $H$ observations from a fixed MDP. The reward function and transition kernel are fixed within each deployment. Across deployments, at most an $\epsilon >0$ fraction of the reward and transition kernel entries may change. In this setting, learning each model to deploy from scratch may discard useful observations from past deployments, whereas pooling all past observations can bias the estimates of entries that have already changed. We then propose an algorithm that for each state and action pair, selects how much history to reuse to estimate the reward and probabilities of the next state. We start with observations from the current deployment and add earlier deployment observations one at a time, the most recent first. After each addition, we check whether a single average reward agrees with the estimates from all histories considered until that point. We apply the same check to the distribution of next states. For each estimate, we leverage the last history that passes these aforementioned checks. The optimal control policy is then computed on top of the fitted MDP. We characterize the \emph{average gap} between the optimal return and the return of the learner's policies over a finite number of deployments.  We demonstrate that when $0<|\mc S|\epsilon\le2$, with $|\mc S|$ being the cardinality of the state space, our algorithm needs only the observations from the most recent $\mathcal O(1/\epsilon)$ deployments, and achieves a \emph{long-run average gap} of at most $\widetilde{\mathcal O}(\sqrt{\epsilon/H})$.
\end{abstract}

\section{Introduction}
\label{sec:introduction}

A control policy learned in one environment may perform poorly when that environment changes. For example, a blocked route changes where a robot can move, while a shift in customer demand can change how much stock a retailer should order \cite{rahimi2023replay,cheung2020reinforcement}.
Such changes need not affect the entire model. If only a few reward or transition entries change, past observations of the remaining entries still provide useful information about the current environment. Learning the underlying deployed model from scratch may discard that information. On the other hand, reusing all past observations can produce inaccurate and biased estimates of the entries that have already changed.
The problem is how to adaptively select which observations to reuse (i.e., how to adapt the memory to the past observations) when the learner does not know where the changes occurred.\vspace{0.1cm}

\begin{figure}[t]
\centering
\hspace{-1mm}\includegraphics[scale=0.25]{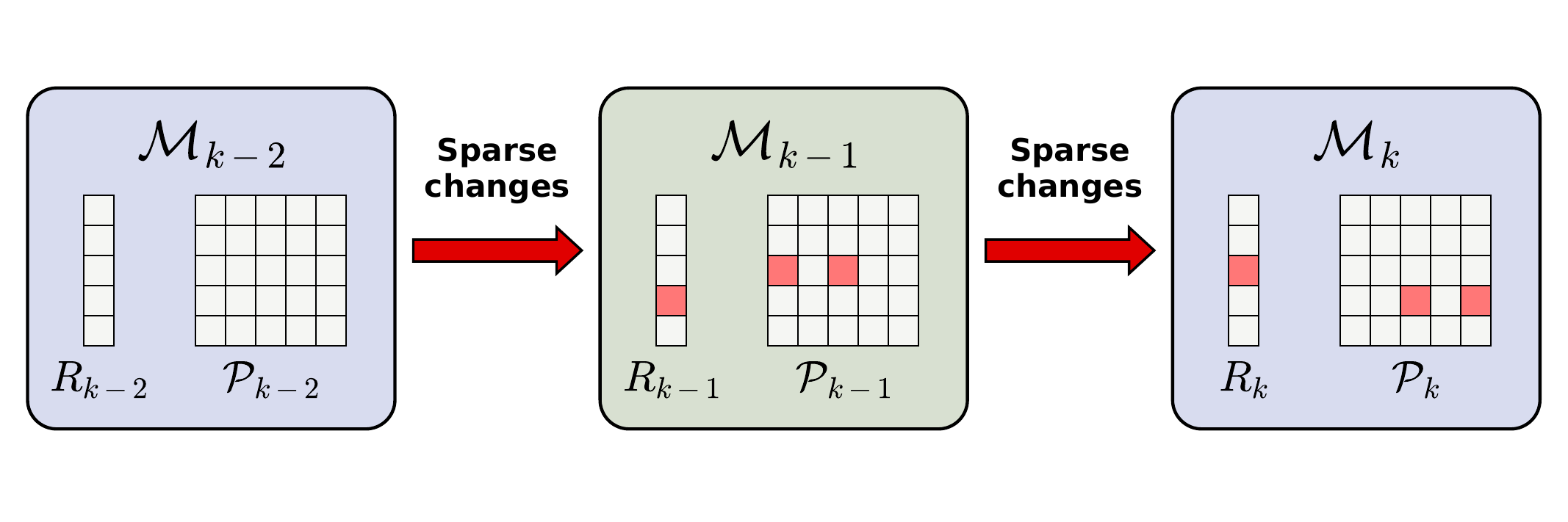}
\vspace{-12 mm}
\caption{\hspace{-5mm}Sparse reward and transition changes across deployments
$\mathcal M_{k-2},\mathcal M_{k-1},\mathcal M_k$.
Red cells are the entries that change.}\vspace{-0.6cm}
\label{fig:sparse-deployments}
\end{figure}

We study discounted infinite-horizon reinforcement learning (RL) across a sequence of deployments. A deployment is a period of $H$
observations during which the reward function and transition kernel are fixed, the deployment boundary is the known time at which the next period begins and the model may change. The state and action spaces are shared across deployments. At each boundary, at most an $\epsilon$ fraction of the reward entries and an $\epsilon$ fraction of the transition-probability entries may change, as illustrated in Fig.~\ref{fig:sparse-deployments}. A change is local in this sense: it affects a small set of the model entries, although it may alter the optimal underlying policy. The learner knows the change budget $\epsilon$ but neither the changed entries nor their magnitudes. 

As observations arrive, the learner updates its policy to track
the changing environment. We measure performance by the average
gap from each deployment's optimal return, including during
adaptation. In particular, we ask the following question: \emph{How much can selective reuse of past observations improve policy tracking under sparse model changes, and how much memory does it require?}

\vspace{0.1cm}
\noindent \textbf{Related Work.} Learning across changing environments involves two closely related goals: adapting to model changes and reusing information from earlier environments. The former is central to nonstationary RL, where existing methods account for model variation~\cite{ortner2020variational}, use sliding windows with confidence widening~\cite{cheung2020reinforcement}, adapt to unknown nonstationarity~\cite{wei2021nonstationary}, or restart after detecting changes~\cite{alami2023restarted}. The latter underlies multitask learning, which exploits shared representations across related tasks~\cite{caruana1997multitask,baxter2000bias,tripuraneni2021linear}, and related control methods that establish sample complexity benefits from sharing information across dynamical systems~\cite{zhang2023multi,toso2024maml,richards2023control,toso2024async}. Bringing adaptation and reuse together, Meta-RL learns policy updates~\cite{alshedivat2018continuous}, dynamics-model priors~\cite{nagabandi2019adapt}, reusable model mixtures~\cite{nagabandi2019deep}, or task representations~\cite{bing2023nonstationary} to facilitate adaptation, while domain randomization trains policies across varied simulator parameters~\cite{tobin2017domain,peng2018dynamics,ramos2019bayessim}. Our work exploits a specific source of shared structure: individual reward entries and transition rows can remain unchanged even when the environment as a whole changes. With known deployment boundaries and sparse model changes, the learner selects a separate history for each component, allowing persistent components to benefit from past observations while adapting estimates of those that change.

Our setting has known deployment boundaries and sparse changes in
reward and transition entries. We exploit \emph{components that remain
unaltered} across deployments by selecting a \emph{separate history for
each reward entry and transition row}. These histories are used to
fit the current model, and we evaluate the resulting policies
against the current deployment's optimal policy throughout adaptation.
A closely related approach uses local forgetting in replay buffers
to adapt to localized reward changes in deep model-based
RL~\cite{rahimi2023replay}. Our results characterize the statistical benefits of a selective history reuse under independent sampling and exact planning. We consider changes in both rewards and transitions and bound the average policy loss from the start of each deployment, including before the learner receives its first new observation. Our contributions are as follows:
\vspace{0.1cm}

\noindent $\bullet$ \textbf{Adaptive history selection.} We propose an algorithm that selects a history separately for each
reward entry and transition row (Section~\ref{sec:algorithm}). For each state–action pair, the learner selects separate histories for estimating the mean reward and the distribution over next states. It starts with the observations from the current deployment and adds earlier deployments one at a time, most recent first. After each addition, it checks whether a single mean reward lies in every confidence interval constructed until that point. It applies the same check to the next-state distribution using the transition confidence sets. For each estimate, it keeps the longest history that passes these checks and computes an optimal policy for the model fitted on top of the selected history.
The algorithm stores only
state-action visit counts, reward sums, and next state counts for each stored deployment. We call this collection a \emph{deployment summary}. \vspace{0.1cm}

\noindent $\bullet$ \textbf{Average policy-loss bounds.}
We establish expected and high-probability bounds over any finite number of deployments (Section~\ref{sec:results}). The bounds account for the initial learning step where the learner has no past observations. In addition, our bounds also accounts for the limit on retained history, and the adaptation to the sparse changes. Our proof controls the
accumulated estimation error across deployments. For
$0<\epsilon\le1$, storing $B=\lceil\epsilon^{-1}\rceil$ completed summaries provides a long-run average policy loss of $\widetilde{\mathcal O}(\sqrt{\epsilon/H})$ in the regime $|\mc S|\epsilon\le2$, where $|\mc S|$ denotes the cardinality of the state space.  The number of completed summaries $B$ is selected to balance the contribution from limited history with respect to the model changes.

\section{Background and Problem Formulation}
\label{sec:formulation}
Consider a discounted MDP $\mc M=(\mc S,\mc A,\mc P,R,\gamma)$
with finite nonempty state and action spaces, transition
kernel $\mc P$, mean reward $R(s,a)=\mathbb E[r\mid s,a]$,
and $\gamma\in(0,1)$. Let $d:=|\mc S||\mc A|$ and let
$\Delta(\mc B)$ denote the probability distributions on a
finite set $\mc B$. Assume $|r|\le\bar R$ almost surely for
a known $\bar R\ge1$. The standard objective is to maximize
$\mathbb E_{\mc P,\pi}[\sum_{t=0}^{\infty}\gamma^tR(s_t,a_t)
\mid s_0=s]$ over stationary policies
$\pi:\mc S\to\Delta(\mc A)$ for every initial state $s$.
A stationary deterministic optimal policy exists~\cite{suttonRL}.

In this paper, we study successive deployments in which the MDP remains
fixed within each deployment but may undergo sparse reward
and transition changes between deployments. Our goal is
\emph{small average discounted policy loss} through selective reuse of past data, without knowing which components have
changed. We formalize this setting next.

\noindent\textcolor{winered}{\(\smallsquare\)}~\textbf{MDPs and Deployment Structure.}
The environment is held fixed during deployments containing
$H\ge1$ observation times each. Deployment $k$, where
$k=0,1,\ldots$, is described by the MDP
\(
 \mc{M}_k=(\mc{S},\mc{A},\mc{P}_k,R_k,\gamma).
 \)
The state space $\mc{S}$, action space $\mc{A}$, and discount
factor $\gamma$ are shared across deployments, while the
transition kernel $\mc{P}_k$ and mean reward function $R_k$
may change between consecutive deployments. For each $(s,a)\in\mc{S}\times\mc{A}$,
$\mc{P}_k(\cdot\mid s,a)\in\Delta(\mc{S})$ specifies the
distribution of the successor state during deployment $k$.
Similarly,
$R_k:\mc{S}\times\mc{A}\to[-\bar R,\bar R]$
specifies the expected reward associated with each
state--action pair. The almost-sure reward bound $|r|\le \bar R$ holds uniformly across deployments, with the same known constant $\bar R$. We index observation times within deployment $k$ by
$h\in\{0,\ldots,H-1\}$ and use $(k,h)$ to identify an
observation time. Thus, $\mc{P}_k$ and $R_k$ remain unchanged
throughout these $H$ observation times, and any change occurs
at a deployment boundary. Both $\mc{P}_k$ and $R_k$ are
unknown to the learner. The learner knows $H$ and the
deployment boundaries, but does not observe which model
entries change or the magnitudes of those changes.
Observations collected in earlier deployments may be
retained and reused when adapting to the current MDP.
We impose sparsity on consecutive changes as follows.

\begin{assumption}\textbf{(Sparse Changes in Model Entries)}
\label{ass:sparse}
For a known $\epsilon\in[0,1]$ and each $k\ge0$, define
\(
 \mc{C}_k^R
 :=\big\{(s,a)\in\mc{S}\times\mc{A}: R_{k+1}(s,a)\ne R_k(s,a)\big\},
 \mc{E}_k^P
 :=\big\{(s,a,s')\in\mc{S}\times\mc{A}\times\mc{S}: \mc{P}_{k+1}(s'\mid s,a)
 \ne\mc{P}_k(s'\mid s,a)\big\}.
\)
At every deployment boundary,
\(
 |\mc{C}_k^R|\le\epsilon d, \hspace{1 mm}
 |\mc{E}_k^P|\le\epsilon|\mc{S}|d.
 \)
\end{assumption}
Assumption~\ref{ass:sparse} bounds $|\mc C_k^R|$ and
$|\mc E_k^P|$. Changes may have any admissible magnitude, and
their locations may vary or recur across boundaries. Let
$\mathcal C_k^P:=\{x:P_{k+1}(\cdot\mid x)\ne P_k(\cdot\mid x)\}$.
Each changed row has at least two changed entries, so
Lemma~\ref{lem:rows} gives $|\mathcal C_k^P|\le\vartheta d$,
where $\vartheta:=\min\{1,|\mc{S}|\epsilon/2\}$.

\noindent\textcolor{winered}{\(\smallsquare\)}~\textbf{Observation Model.}
At observation time $h\in\{0,\ldots,H-1\}$ of deployment $k$,
the learner receives a tuple
\begin{equation}
\begin{aligned}
(s_{k,h},a_{k,h},r_{k,h},s'_{k,h})\in
 \mc{S}\times\mc{A}\times[-\bar R,\bar R]\times\mc{S}.
\end{aligned}
\label{eq:tuple}
\end{equation}
Here, $(s_{k,h},a_{k,h})$ is the sampled state--action pair,
$r_{k,h}$ is the observed reward, and $s'_{k,h}$ is the
sampled successor state. We use the following independent
sampling model.

\begin{assumption}\textbf{(Observation Tuples)}
\label{ass:data}
The environment sequence is fixed independently of the observation
noise. Conditional on this sequence, the tuples are independent
across $(k,h)$, and $(s_{k,h},a_{k,h})$ is drawn from $\nu_k$, where $\nu_k(s,a)\ge v>0$ for every $(s,a)\in\mc{S}\times\mc{A}$. Given $(s_{k,h},a_{k,h})=(s,a)$, the reward $r_{k,h}$ has mean $R_k(s,a)$ and the successor state $s'_{k,h}$ has distribution
$\mc{P}_k(\cdot\mid s,a)$. The constants $v$ and $\bar R$ are known. The reward and next state within a tuple may be dependent.
\end{assumption}
The sampling distributions $\nu_k$ may vary across deployments
and be unknown, with $v\le 1/d$. The bounds hold for fixed
environment sequences and conditionally for exogenous random
ones. Under Assumption~\ref{ass:data}, overlapping histories
may be dependent because they share tuples. Before observing
tuple $(k,h)$, the learner computes $\pi_{k,h}$ from retained
past data and the first $h$ tuples of deployment $k$.
\subsection{Policy Values and Discounted Occupancies}
\label{sub:occupancy}
Fix an initial distribution $\rho\in\Delta(\mc S)$ common
to all deployments. For a stationary policy $\pi$, define
$V_k^\pi(s):=\mathbb E_{\mc P_k,\pi}
[\sum_{t=0}^{\infty}\gamma^tR_k(s_t,a_t)\mid s_0=s]$,
where trajectories evolve under $\pi$ in the fixed MDP
$\mc M_k$. The expected return from $\rho$ and its optimal
value are $J_k^\rho(\pi):=\rho^\top V_k^\pi$ and
$J_k^{\rho,*}:=\max_\pi J_k^\rho(\pi)$, respectively.
Since $\mc M_k$ is finite and discounted, a stationary
deterministic optimal policy $\pi_k^*$ exists and satisfies
$J_k^\rho(\pi_k^*)=J_k^{\rho,*}$.
To quantify visitation under $\pi$ starting from $s_0\sim\rho$,
define the normalized discounted occupancy
$d_{\rho,k}^\pi(s,a):=(1-\gamma)\sum_{t=0}^{\infty}\gamma^t
\mathbb P_{\rho,\mc P_k,\pi}(s_t=s,a_t=a)$.
These visitation weights satisfy
$\sum_{s,a}d_{\rho,k}^\pi(s,a)=1$ and hence form a
probability distribution on $\mc S\times\mc A$.
\begin{assumption}\textbf{(Occupancy Bound)}
\label{ass:occupancy}
There exists a known $\kappa\in[1,d]$ such that, for every
deployment $k$, stationary policy $\pi$, and
$(s,a)\in\mc{S}\times\mc{A}$,
\(
 d_{\rho,k}^\pi(s,a)\le\frac{\kappa}{d}.
 \)
\end{assumption}

Assumption~\ref{ass:occupancy} bounds discounted policy
occupancies under $\mc P_k$, separately from the sampling
coverage in Assumption~\ref{ass:data}. Since occupancies are
probability distributions, $\kappa=d$ is always valid;
smaller $\kappa$ improves the guarantees under sparse changes. Define the policy loss in deployment $k$ by
$\ell_k^\rho(\pi):=J_k^{\rho,*}-J_k^\rho(\pi)$.
The reward bound gives
$0\le\ell_k^\rho(\pi)\le2\bar R/(1-\gamma)$. For $K\ge1$ deployments, define the average policy loss and
its expectation by
\begin{equation}
 \mc{L}_{K,H}^\rho :=\frac1{KH}\sum_{k=0}^{K-1}\sum_{h=0}^{H-1}
 \ell_k^\rho(\pi_{k,h}),\hspace{1 mm} \mc{R}_{K,H}^\rho :=\mathbb E[\mc{L}_{K,H}^\rho].
 \label{eq:loss}
\end{equation}
Each summand evaluates $\pi_{k,h}$ in the fixed MDP
$\mc{M}_k$ from the common initial distribution $\rho$.
Thus, the average includes the policy available at the
start of each deployment and every policy computed during
adaptation. The expectation is over the observation
randomness and any randomness in the environment sequence.\\[1 mm]
\noindent\fbox{%
\begin{minipage}{\dimexpr\linewidth-2\fboxsep-2\fboxrule\relax}
\begin{problem}
\label{prob:tracking}
Suppose Assumptions~\ref{ass:sparse}--\ref{ass:occupancy} hold. Construct a causal learner that reuses past observations
without knowing which model entries change. Bound
$\mc{R}_{K,H}^\rho$ for every $K\ge1$. For a fixed learner
and deployment length $H$, also bound
 \(J_H^\rho:=
 \limsup_{K\to\infty}\mc{R}_{K,H}^\rho,\)
with explicit dependence on the entry-change fraction
$\epsilon$.
\end{problem}
\end{minipage}%
}
\section{Our Proposed Algorithm}
\label{sec:algorithm}
In this section, we develop an adaptive history estimator
for reusing data across deployments. A change to one reward
entry or transition row need not invalidate past data for
other components. A short common history discards \emph{useful
observations for unchanged components}, while a long one can
\emph{bias estimates of changed components}. We therefore select
histories \emph{separately} for each reward entry and transition row. To choose histories without knowing change times, we use the fact that estimates of an unchanged component should agree up to sampling noise. The confidence sets quantify this uncertainty, and we extend histories while their intersection remains nonempty. The learner then fits a valid MDP within the selected intersections and computes a stationary optimal policy before the next observation. We first define the
candidate histories and their empirical estimates.

\noindent \textcolor{winered}{\(\smallsquare\)}~\textbf{Candidate Histories and Empirical Estimates.}
Fix an integer history cap $B\ge1$ and a confidence parameter
$\delta\in(0,1)$. The cap $B$ limits the number of completed
deployments available for reuse. Define
\begin{equation}
 L:=\log\frac{8|\mc{S}||\mc{A}|(B+1)(H+1)}{\delta},
 \hspace{1 mm} \Lambda:=|\mc{S}|\log2+L.
 \label{eq:logs}
\end{equation}
The factor $L$ accounts for simultaneous confidence bounds
over histories, state--action pairs, and observation times;
$\Lambda$ additionally accounts for the dimension of transition
rows. At $(k,h)$, write
\((\textcolor{winered}{\bullet})\triangleq
{\bigcup}_{j=k-m}^{k-1}\{(j,u):0\le u<H\}\)
for the indices from the last $m$ completed deployments.
Including the first $h$ observation indices from deployment
$k$ gives the candidate history as follows:
\begin{equation}
\begin{aligned}
 \mc{W}_{k,h}^{(m)}
 &:= (\textcolor{winered}{\bullet}) \cup\{(k,u):0\le u<h\},
\end{aligned}
\label{eq:window}
\end{equation}
where $0\le m\le\min\{B,k\}$. All included observations
precede $(k,h)$, with $m=0$ using only data from deployment
$k$. For each $(s,a)\in\mc S\times\mc A$, define the visit count
\begin{equation}\label{eq:visit-counts}
N_{k,h}^{(m)}(s,a):=\sum_{(j,u)\in\mc W_{k,h}^{(m)}}
\mathbf{1}\{(s_{j,u},a_{j,u})=(s,a)\}.
\end{equation}
This count determines the number of observations contributing
to the estimates for $(s,a)$. For $N_{k,h}^{(m)}(s,a)>0$, define the empirical reward mean
\begin{equation}
\overline R_{k,h}^{(m)}(s,a):=\frac{
 \sum_{(j,u)\in\mc W_{k,h}^{(m)}}
 \mathbf{1}\{(s_{j,u},a_{j,u})=(s,a)\}r_{j,u}}
 {N_{k,h}^{(m)}(s,a)},
\label{eq:emp-reward}
\end{equation}
and, for each $s'\in\mc S$, the empirical transition probability
\begin{equation}
\overline{\mc P}_{k,h}^{(m)}(s'\mid s,a):=\frac{
 \sum_{(j,u)\in\mc W_{k,h}^{(m)}}
 \mathbf{1}\{(s_{j,u},a_{j,u},s'_{j,u})=(s,a,s')\}}
 {N_{k,h}^{(m)}(s,a)}.
\label{eq:emp-transition}
\end{equation}
For $N_{k,h}^{(m)}(s,a)>0$, define the confidence radii
$b_{k,h}^{R,(m)}(s,a):=\bar R(2L/N_{k,h}^{(m)}(s,a))^{1/2}$
and $b_{k,h}^{P,(m)}(s,a):=(2\Lambda/N_{k,h}^{(m)}(s,a))^{1/2}$.
The reward radius controls the absolute deviation of the
empirical mean from $R_k(s,a)$, while the transition radius
controls the $\ell_1$ deviation of the empirical row from
$\mc P_k(\cdot\mid s,a)$. These guarantees hold with high
probability when the corresponding component remains unchanged
throughout the candidate history. Now, we define the reward
confidence interval as
\begin{equation}\label{eq:reward-interval}
\medmath{
 I_{k,h}^{R,(m)}(s,a) :=\big\{z\in[-\bar R,\bar R]: |z-\overline R_{k,h}^{(m)}(s,a)|
 \le b_{k,h}^{R,(m)}(s,a)\big\}.}
\end{equation}
As in \eqref{eq:reward-interval}, we define the transition
confidence set as
\begin{equation}\medmath{
\begin{aligned}
 \mc{B}_{k,h}^{P,(m)}(s,a):=\big\{p\in\Delta(\mc{S}): \norm*{p-\overline{\mc{P}}_{k,h}^{(m)}(\cdot\mid s,a)}_1
 \le b_{k,h}^{P,(m)}(s,a)\big\}.
\end{aligned}}
\label{eq:transition-ball}
\end{equation}
For $N_{k,h}^{(m)}(s,a)=0$, set
$I_{k,h}^{R,(m)}(s,a)=[-\bar R,\bar R]$ and
$\mc B_{k,h}^{P,(m)}(s,a)=\Delta(\mc S)$,
so the history imposes only the known reward range and
transition stochasticity.
\vspace{0.1cm}
\begin{algorithm}[h]
\caption{Adaptive History Model Learning}
\label{alg:tracking}
\begin{algorithmic}[1]
\Require Observations per deployment $H$; maximum number of retained
past deployments $B$; confidence parameter $\delta$; discount factor $\gamma$; state and action spaces
$\mc S,\mc A$.
\State Compute $L,\Lambda$ using \eqref{eq:logs}.
\For{$k=0,1,\ldots$}
 \State Reset the current-deployment visit counts
\eqref{eq:visit-counts}, and the reward sums and next-state counts
used in \eqref{eq:emp-reward}--\eqref{eq:emp-transition}.
 \For{$h=0,\ldots,H-1$}
  \State Construct candidate confidence sets for all
  $(s,a)\in\mc S\times\mc A$ and $m=0,\ldots,\min\{B,k\}$
  using \eqref{eq:emp-reward}--\eqref{eq:transition-ball},
  with the zero-count convention.
  \State Form the confidence intersections using \eqref{eq:intersection}.
  \State Select the history indices using
  \eqref{eq:select-reward}--\eqref{eq:select-transition}.
  \State Construct $\widehat R_{k,h},\widehat{\mc P}_{k,h}$
  using \eqref{eq:reward-estimate}--\eqref{eq:transition-estimate}.
  \State Output stationary optimal policy $\pi_{k,h}$
  using \eqref{eq:planning}.
  \State Receive data tuple \eqref{eq:tuple} and update statistics.
 \EndFor
 \State Store the current summary and keep the latest $B$.
\EndFor
\end{algorithmic}
\end{algorithm}

\noindent \textcolor{winered}{$\smallsquare$}~\textbf{Adaptive History Selection.}
The reward mean and transition row at $(s,a) \in \mc{S} \times \mc{A}$ may change at
different boundaries, so we select their histories separately
by intersecting confidence sets over nested windows:
\begin{align}\label{eq:intersection}\medmath{
 J_{k,h}^{R,(m)}(s,a) :=\bigcap_{\ell=0}^{m}I_{k,h}^{R,(\ell)}(s,a),\hspace{1 mm}
 \mc{J}_{k,h}^{P,(m)}(s,a) :=\bigcap_{\ell=0}^{m}\mc{B}_{k,h}^{P,(\ell)}(s,a).}
\end{align}
A nonempty intersection admits a common reward mean or
transition row satisfying all constraints up to $m$.
As the intersections shrink with $m$, we select the
largest indices for which they remain nonempty:
\begin{align}
 \widehat m_{k,h}^R(s,a)
 &:=\max\{m:J_{k,h}^{R,(m)}(s,a)\ne\varnothing\},
 \label{eq:select-reward}\\
 \widehat m_{k,h}^P(s,a)
 &:=\max\{m:\mc{J}_{k,h}^{P,(m)}(s,a)\ne\varnothing\},
 \label{eq:select-transition}
\end{align}
where both maxima range over $0\le m\le\min\{B,k\}$.
The reward estimate is the midpoint of its selected interval:
\begin{equation}
 \widehat R_{k,h}(s,a)
 :=\mathbf{mid}
 \left[J_{k,h}^{R,(\widehat m_{k,h}^R(s,a))}(s,a)\right].
 \label{eq:reward-estimate}
\end{equation}
This choice satisfies every reward confidence constraint in
the selected prefix. For transitions, let $\mathbf u$ be the
uniform distribution on $\mc{S}$ and write
\((\textcolor{myblue}{\bullet})\triangleq
\mc{J}_{k,h}^{P,(\widehat m_{k,h}^P(s,a))}(s,a)\)
for the selected confidence intersection. Define
\begin{equation}
 \widehat{\mc{P}}_{k,h}(\cdot\mid s,a):=
 \argmin_{p\in(\textcolor{myblue}{\bullet})}
 \norm{p-\mathbf u}_2^2.
 \label{eq:transition-estimate}
\end{equation}
The nonempty $m=0$ sets ensure that both selected indices
are well defined. The reward intersection is a compact
interval, and the transition intersection is compact and
convex, so the projection in \eqref{eq:transition-estimate}
is unique. Both estimates are measurable functions of
observations available at $(k,h)$. Selection uses only
confidence-set consistency and requires no knowledge of
change locations or magnitudes.

\vspace{0.1cm}
\noindent \textcolor{winered}{$\smallsquare$}~\textbf{Computing the Policy.} The estimates $\widehat R_{k,h}$ and
$\widehat{\mc{P}}_{k,h}$ define a fitted MDP at time $(k,h)$.
To compute an optimal policy for this model, solve the
Bellman optimality equation
\begin{equation}
\begin{aligned}
 \widehat V_{k,h}(s)
 &=\max_{a\in\mc{A}}\bigg\{\widehat R_{k,h}(s,a) +\gamma\sum_{s'\in\mc{S}}
 \widehat{\mc{P}}_{k,h}(s'\mid s,a)
 \widehat V_{k,h}(s')\bigg\}.
\end{aligned}
\label{eq:planning}
\end{equation}
The fitted Bellman optimality operator is a $\gamma$-contraction
with unique fixed point $\widehat V_{k,h}$. Solving \eqref{eq:planning} yields a stationary deterministic optimal policy $\pi_{k,h}$ for the fitted MDP.This policy is computed before receiving tuple $(k,h)$, using retained summaries and
the first $h$ observations of deployment $k$. The newly received
tuple then updates the current statistics.
Algorithm~\ref{alg:tracking} summarizes the procedure.

Algorithm~\ref{alg:tracking} retains visit counts, reward sums,
and successor counts for the latest $B$ completed deployments
and the current one. These summaries suffice for all candidate
estimates and require $\mathcal O((B+1)|\mc S|d)$ storage,
without retaining individual tuples.
\section{Main Results}
\label{sec:results}
In this section, we present our main results for
Algorithm~\ref{alg:tracking}. We recall the average policy loss $\mc{L}_{K,H}^\rho$ and its expectation $\mc{R}_{K,H}^\rho$, as defined in \eqref{eq:loss}. In Theorem~\ref{thm:main}, we bound the expected
average policy loss over finitely many deployments, and
in Corollary~\ref{cor:hp} we provide the corresponding high-probability bound. Moreover, Theorem~\ref{cor:persistent} establishes a long-run
tracking guarantee while keeping only
$\mathcal O(\epsilon^{-1})$ deployment summaries.

\begin{theorem}\textbf{(Finite-deployment policy loss)}
\label{thm:main}
Suppose Assumptions~\ref{ass:sparse}-\ref{ass:occupancy} hold. Then, for every $K\ge1$, Algorithm~\ref{alg:tracking}, with an integer
$B\ge1$ and $\delta\in(0,1)$, satisfies
\begin{equation}\label{eq:main-bound}
\medmath{
\begin{aligned}
 \hspace{-1mm}\mc{R}_{K,H}^\rho
 \lesssim
 \frac{\bar R}{(1-\gamma)^2}\left[
 \sqrt{\frac{L}{vH}
 \left(\bullet+\kappa\epsilon\right)}+
 \sqrt{\frac{\Lambda}{vH}
 \left(\bullet+\kappa\vartheta\right)}\right]+\frac{\bar R\delta}{1-\gamma},
\end{aligned}
}
\end{equation}
where $L,\Lambda$ are defined in \eqref{eq:logs},
$(\bullet):=B^{-1}+K^{-1}$, and $\vartheta:=\min\{1,|\mc{S}|\epsilon/2\}$.
Here, $K$ is the number of deployments, $H$ is the number of observations per deployment, $B$ is the maximum number of past deployments retained for estimation, and $v>0$ is the uniform lower bound on the sampling probabilities
$\nu_k(s,a)$ in Assumption~\ref{ass:data}.
\end{theorem}
\noindent \textbf{Discussion of Theorem~\ref{thm:main}.}
The bound in \eqref{eq:main-bound} separates the effects of
initial learning, finite-history estimation, and adaptation
to sparse model changes. The first two contributions decay
as $K^{-1/2}$ and $B^{-1/2}$, respectively, and both scale
as $(vH)^{-1/2}$ up to logarithmic factors. Sparse changes
enter through $\sqrt{\kappa\epsilon}$ for rewards and
$\sqrt{\kappa\vartheta}$ for transitions, where
$\vartheta=\min\{1,|\mc{S}|\epsilon/2\}$ makes the dependence
on the number of states explicit. Intuitively, selective
history reuse allows unchanged components to benefit from
past observations while limiting the influence of outdated
data. For fixed $B$ and $\delta$, the confidence factors
are independent of $K$. 

\begin{corollary}\textbf{(High-probability policy loss)}
\label{cor:hp}
Under the conditions of Theorem~\ref{thm:main}, fix $K\ge1$
and $\alpha\in(0,1)$. Run Algorithm~\ref{alg:tracking} with
$\delta=\alpha/K$, and define
$L_\alpha:=\log(8dK(B+1)(H+1)/\alpha)$ and
$\Lambda_\alpha:=|\mc{S}|\log2+L_\alpha$.
With probability at least $1-\alpha$, the following bound holds:
\begin{equation}\medmath{
\begin{aligned}
 \mc{L}_{K,H}^\rho
 \lesssim \frac{\bar R}{1-\gamma}
 \sqrt{\frac{L_\alpha(\bullet+\kappa\epsilon)}{vH}} +\frac{\gamma\bar R}{(1-\gamma)^2}
 \sqrt{\frac{\Lambda_\alpha(\bullet+\kappa\vartheta)}{vH}},
\end{aligned}}
\label{eq:hp-bound}
\end{equation}
where $(\bullet)$, and all remaining quantities defined in Theorem~\ref{thm:main}.
\end{corollary}
Lastly, we study persistent tracking with a learner
whose parameters are independent of $K$. The following
theorem shows that retaining $\mathcal O(\epsilon^{-1})$
deployment summaries suffices to control the long-run
expected average policy loss.
\begin{theorem}\textbf{(Persistent Tracking)}
\label{cor:persistent}
Under the conditions of Theorem~\ref{thm:main}, let
$0<\epsilon\le1$. Choose $B=\ceil{\epsilon^{-1}}$ and
$\delta=\epsilon/(H+1)$, and define
$L_\epsilon:=\log(24d(H+1)^2/\epsilon^2)$ and
$\Lambda_\epsilon:=|\mc{S}|\log2+L_\epsilon$.
For every $K\ge1$, the same learner satisfies
\begin{equation}\medmath{
\begin{aligned}
 \mc{R}_{K,H}^\rho
 \lesssim \frac{\bar R}{1-\gamma}
 \sqrt{\frac{L_\epsilon(K^{-1}+\kappa\epsilon)}{vH}} +\frac{\gamma\bar R}{(1-\gamma)^2}
 \sqrt{\frac{\Lambda_\epsilon(K^{-1}+\kappa\vartheta)}{vH}}.
\end{aligned}}
\label{eq:persistent-finite}
\end{equation}
Therefore, we have
\begin{equation}
 J_H^\rho
 \lesssim 
 \frac{\bar R}{1-\gamma}\sqrt{\frac{\kappa}{vH}}
 \left[
 \sqrt{\epsilon L_\epsilon}
 +\frac{\gamma\sqrt{\vartheta\Lambda_\epsilon}}{1-\gamma}
 \right].
\label{eq:persistent-longrun}
\end{equation}
\end{theorem}
\noindent \textbf{Discussion of Theorem~\ref{cor:persistent}.} In \eqref{eq:main-bound}, choosing
$B=\ceil{\epsilon^{-1}}$ gives $B^{-1}\le\epsilon$,
so the $B^{-1}$ contributions are absorbed, up to absolute
constants, into the terms involving $\kappa\epsilon$ and
$\kappa\vartheta$. Choosing $\delta=\epsilon/(H+1)$
makes the final term
$\bar R\epsilon/((1-\gamma)(H+1))$, which is dominated
by the reward-change contribution. As neither choice
depends on $K$, the same learner operates across arbitrarily
many deployments while retaining only
$\mathcal O(\epsilon^{-1})$ deployment summaries.
As $K\to\infty$, the $K^{-1}$ contributions vanish.
When $|\mc{S}|\epsilon\le2$, substituting
$\vartheta=|\mc{S}|\epsilon/2$ into
\eqref{eq:persistent-longrun} yields
$J_H^\rho=\widetilde{\mathcal O}(\sqrt{\epsilon/H})$
for fixed remaining problem parameters.
\section{Numerical Experiments}
\label{sec:experiments}
We consider an MDP with $|\mc S|=50$, $|\mc A|=4$, $\gamma=0.8$, and uniform initial distribution $\rho$. Each state-action pair has two fixed successors with probabilities $(p,1-p)$, where $p\sim\mathrm{Unif}[0.15,0.85]$. The rewards take values
in $\{-1,1\}$, with initial means uniform on $[-0.10,0.10]$. In addition, reward and transition changes occur within separate, fixed subsets, each containing $20\%$ of the state-action pairs and they are unknown to the learner. At each boundary, $200\epsilon$
reward entries and $200\epsilon$ transition rows are selected from their respective subsets and resampled. Replacement reward means are uniform on $[-0.65,0.65]$, and transition probabilities are sampled as above. We consider $\epsilon\in\{0.02,0.005\}$.

We compare against \emph{restart}, using only current
deployment data, \emph{full history}, pooling all available
observations, and \emph{shared history}, using the shortest
history selected across all components as a common window
for the entire MDP. The proposed method selects histories
separately for each reward entry and transition row.
All methods use identical observations and compute optimal policies in their fitted MDPs at common, geometrically spaced sample counts within each deployment. Each policy's loss in the true MDP is weighted by the number of observations for which it is held. Results average $50$ independent runs.
\begin{figure}[ht]
\begin{center}
\includegraphics[scale=0.38]{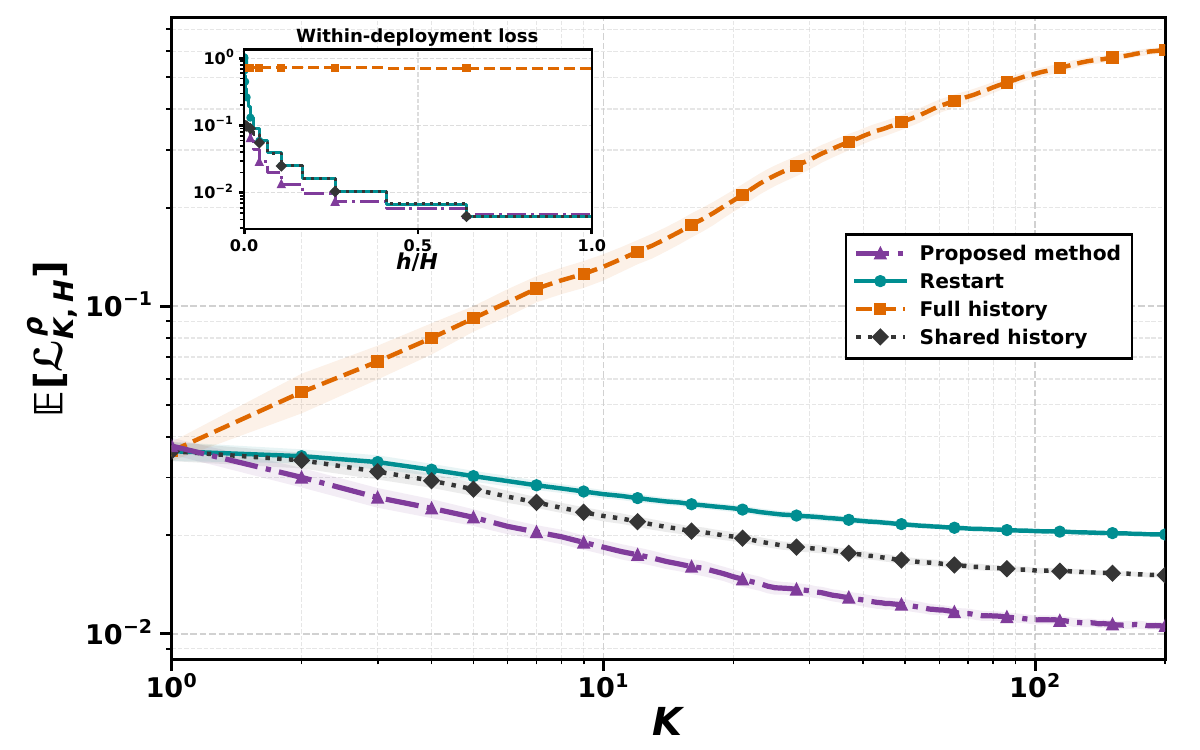}
\includegraphics[scale=0.38]{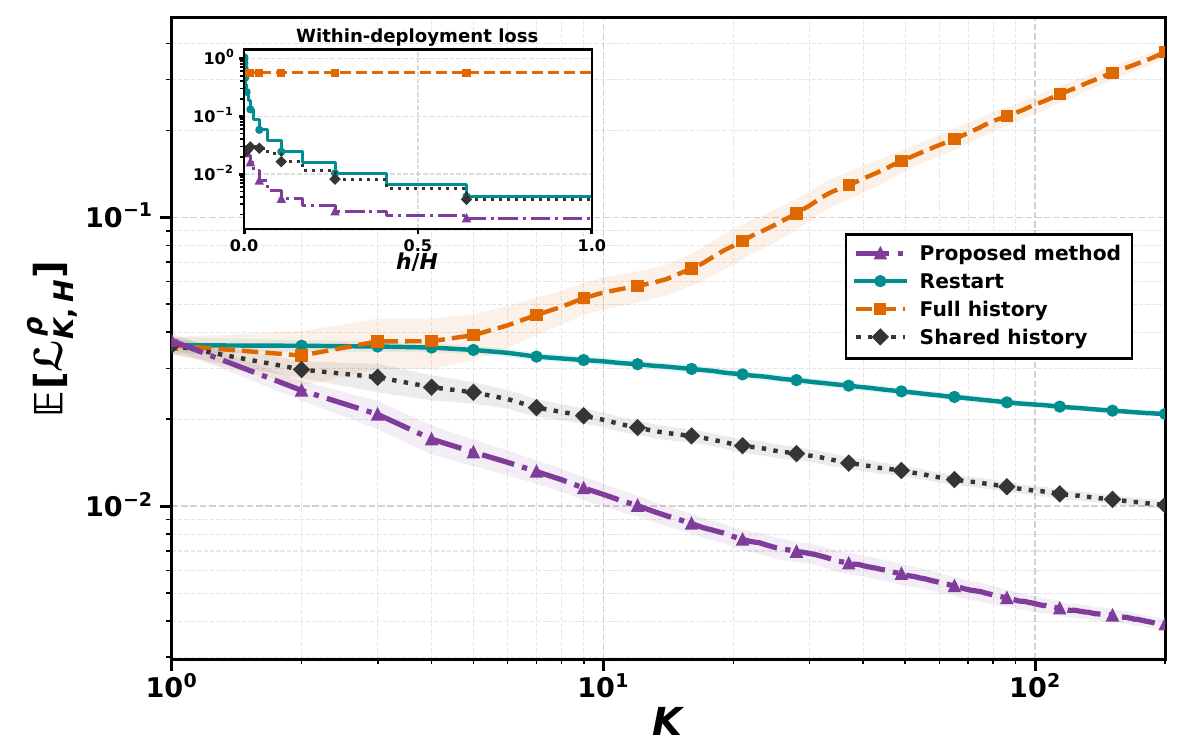} 
\end{center}
\caption{\textbf{(Top)} Evolution of the expected average
policy loss $\mathbb E[\mc L_{K,H}^{\rho}]$ for
$\epsilon=0.02$, with four reward entries and four transition rows changing per deployment boundary.
\textbf{(Bottom)} Results for $\epsilon=0.005$, with one reward entry and one transition row changing per boundary. In both settings, the proposed method achieves the lowest average policy loss by the end of the experiment, while the loss of \emph{full history} increases as deployments accumulate. Over the
final $40$ deployments, our method reduces average policy loss relative to \emph{shared history} by $28.2\%$ for $\epsilon=0.02$ and $65.2\%$ for $\epsilon=0.005$. The insets show that reusing past observations also reduces the loss relative to \emph{restart} early in each deployment.}
\label{fig:sim-1}
\end{figure}
\section{Proof of Main Results}
\label{sec:proofs}
We condition on an arbitrary admissible environment sequence
under Assumption~\ref{ass:data}. Lemma~\ref{lem:rows} converts
the entrywise budget into a bound on changed transition rows.
Lemmas~\ref{lem:concentration} and~\ref{lem:adaptive} use
confidence coverage on unchanged histories to bound the errors
of the selected estimates. Lemma~\ref{lem:accumulation} controls
their weighted average in terms of the total number of component
changes, allowing data-dependent weights. Lemma~\ref{lem:comparison}
then yields the policy loss bound in Theorem~\ref{thm:main}.
All bounds are uniform over admissible sequences.
\subsection{From the Transition Entries to the Affected Rows}
The following lemma converts the entrywise budget in
Assumption~\ref{ass:sparse} into a bound on affected transition rows.
\begin{lemma}\textbf{(Affected transition rows)}
\label{lem:rows}
Under Assumption~\ref{ass:sparse}, the number of affected
transition rows at each boundary $k$ satisfies
$|\mathcal C_k^P|\le|\mathcal E_k^P|/2$ and, therefore,
$|\mathcal C_k^P|\le\vartheta d$.
\end{lemma}
\begin{proof}
For each $(s,a)\in\mc C_k^P$, the difference
$\mc P_{k+1}(\cdot\mid s,a)-\mc P_k(\cdot\mid s,a)$ is nonzero
and sums to zero, so it has both a positive and a negative entry.
Since each changed entry belongs to a unique row,
$2|\mc C_k^P|\le|\mc E_k^P|\le\epsilon|\mc S|d$.
Combining this with $|\mc C_k^P|\le d$ proves Lemma~\ref{lem:rows}.
\end{proof}
With the bound from Lemma~\ref{lem:rows} on the number of transition rows that can change at each deployment boundary, we next turn to the statistical guarantees. Lemma~\ref{lem:concentration} establishes simultaneous confidence coverage within each deployment and lower bounds on visit counts in sufficiently long windows.
\subsection{Concentration and Adaptive History Estimation}
For each $(s,a)\in\mc{S}\times\mc{A}$ and each candidate window \eqref{eq:window}, the reward coverage guarantee in \eqref{eq:reward-interval} applies if the reward mean equals its current value throughout the window. The transition coverage guarantee in \eqref{eq:transition-ball} applies under the analogous condition for the transition row. These conditions depend only on the underlying MDP sequence, hold vacuously for empty windows, and are used only in the analysis.
\begin{lemma}\textbf{(Concentration within a deployment)}
\label{lem:concentration}
For each deployment $k$, define
\(
\mc G_k:=\mc G_k^R\cap\mc G_k^P\cap\mc G_k^N,
\)
where the three events are specified below as follows:

\begin{itemize}
\item On $\mc G_k^R$, $R_k(s,a)\in I_{k,h}^{R,(m)}(s,a)$
whenever $R_j(s,a)=R_k(s,a)$ for all
$(j,u)\in\mc W_{k,h}^{(m)}$.

\item On $\mc G_k^P$,
$\mc P_k(\cdot\mid s,a)\in\mc B_{k,h}^{P,(m)}(s,a)$
whenever $\mc P_j(\cdot\mid s,a)=\mc P_k(\cdot\mid s,a)$
for all $(j,u)\in\mc W_{k,h}^{(m)}$.

\item On $\mc G_k^N$, $N_{k,h}^{(m)}(s,a)\ge \frac{v(mH+h)}{2}$
for $mH+h\ge8L/v$.
\end{itemize}
 All requirements
hold simultaneously over $(s,a)\in\mc S\times\mc A$,
$0\le h\le H$, and $0\le m\le\min\{B,k\}$. Then $\mathbb P(\mc G_k)\ge1-\delta$.
\end{lemma}
\begin{proof}
Fix a candidate history $\mc W_{k,h}^{(m)}$ from
\eqref{eq:window} and a state--action pair $(s,a) \in \mc{S} \times \mc{A}$.
Condition on the sampled state--action pairs in this history,
which fixes the count $N:=N_{k,h}^{(m)}(s,a)$.
First suppose $N>0$. If $R_j(s,a)=R_k(s,a)$ for every
$(j,u)\in\mc W_{k,h}^{(m)}$, then Assumption~\ref{ass:data}
ensures that the $N$ rewards averaged in
\eqref{eq:emp-reward} are conditionally independent with
common mean $R_k(s,a)$. As each reward lies in
$[-\bar R,\bar R]$, Hoeffding's
inequality~\cite{chung2006concentration} yields the following:
\begin{equation}
 \mathbb P\!\left(
 |\overline R_{k,h}^{(m)}(s,a)-R_k(s,a)|
 >\bar R\sqrt{\frac{2L}{N}}\right)
 \le2e^{-L}.
 \label{eq:reward-hoeffding}
\end{equation}
As $R_k(s,a)\in[-\bar R,\bar R]$, intersecting the confidence interval with this range preserves the coverage.

For transitions, suppose $\mc P_j(\cdot\mid s,a)=p:=\mc P_k(\cdot\mid s,a)$ for every $(j,u)\in\mc W_{k,h}^{(m)}$, and write $\overline p:=\overline{\mc P}_{k,h}^{(m)}(\cdot\mid s,a)$
for the empirical row in \eqref{eq:emp-transition}.
Therefore, conditional on all sampled pairs $(s_{j,u},a_{j,u})$ with
$(j,u)\in\mc W_{k,h}^{(m)}$, Assumption~\ref{ass:data}
ensures that the $N$ next states corresponding to
visits to $(s,a)$ are independent with common distribution $p$. For each fixed vector $\sigma\in\{-1,1\}^{|\mc S|}$, let $\sigma(s')$ denote its coordinate corresponding to state $s'\in\mc S$. The $N$ transformed observations $\sigma(s'_{j,u})$ are independent, lie in $[-1,1]$, and have common mean $\sigma^\top p$. Their empirical mean is  then $\sigma^\top\overline p$.
Hence, one-sided Hoeffding inequality ~\cite{chung2006concentration}
yields $\mathbb P(\sigma^\top(\overline p-p)>b)
\le e^{-Nb^2/2}$ for every $b>0$. Moreover, we obtain
$\norm{\overline p-p}_1
=\max_{\sigma\in\{-1,1\}^{|\mc S|}}
\sigma^\top(\overline p-p)$, with the maximum attained by
selecting $\sigma(s')=-1$ when $\overline p(s')<p(s')$
and $\sigma(s')=1$ otherwise. A union bound over the
$2^{|\mc S|}$ sign vectors, with $b=\sqrt{2\Lambda/N}$, yields
\begin{equation}
 \mathbb P\!\left(
 \norm{\overline p-p}_1>\sqrt{\frac{2\Lambda}{N}}\right)
 \le2^{|\mc S|}e^{-\Lambda}=e^{-L}.
 \label{eq:row-concentration}
\end{equation}
As $p\in\Delta(\mc S)$, \eqref{eq:row-concentration} gives
coverage by the set in \eqref{eq:transition-ball}. For $N=0$, both coverage claims hold by construction. Averaging over the sampled state--action pairs gives the same unconditional failure bounds. For the visit-count bound, set $q:=mH+h$. Thus, by Assumption~\ref{ass:data}, $N$ is a sum of $q$ independent Bernoulli indicators with mean $\mu:=\mathbb E[N] =\sum_{(j,u)\in\mc W_{k,h}^{(m)}}\nu_j(s,a)\ge vq$.
The Chernoff inequality \cite[Theorem~3.2]{chung2006concentration} then provides the following bound
\begin{equation}
 \mathbb P(N<vq/2)
 \le\mathbb P(N<\mu/2)
 \le e^{-\mu/8}\le e^{-vq/8}.
 \label{eq:count-chernoff}
\end{equation}
Hence, the count bound fails with probability at most $e^{-L}$ whenever $q\ge8L/v$. Together with the reward and transition bounds, this yields a combined failure probability of at most $4e^{-L}$ for each admissible $((s,a) \in \mc{S} \times \mc{A},m,h)$. 

As $\mc G_k$ requires these bounds simultaneously over at most $d(B+1)(H+1)$ index choices, a union bound and \eqref{eq:logs}
yield $\mathbb P(\mc G_k^c)\le4d(B+1)(H+1)e^{-L}
=\delta/2\le\delta$.
\end{proof}
We use Lemma~\ref{lem:concentration} to bound estimation errors
in terms of component age. For $j\in\{R,P\}$, let
$\tau_k^j(s,a)$ be the largest $m\in\{0,\ldots,k\}$ for which
component $j$ at $(s,a)$ is constant over deployments
$k-m,\ldots,k$. Its age before $(k,h)$ is
$q_{k,h}^j(s,a):=\tau_k^j(s,a)H+h$ observation times.
Define $e_{k,h}^R(s,a):=|\widehat R_{k,h}(s,a)-R_k(s,a)|$
and $e_{k,h}^P(s,a):=\norm*{\widehat{\mc P}_{k,h}(\cdot\mid s,a)
-\mc P_k(\cdot\mid s,a)}_1$.
The history lengths are used only in the analysis and need
not be known to the learner.
\begin{lemma}\textbf{(Adaptive History Error)}
\label{lem:adaptive}
On $\mc G_k$, the following bounds hold for every
$(s,a)\in\mc S\times\mc A$ and $0\le h<H$:
\begin{equation}\medmath{
 \hspace{-1mm}e_{k,h}^R(s,a)
 \le6\bar R\min\left\{1,
 \sqrt{\frac{L}
 {v(\min\{B,\tau_k^R(s,a)\}H+h+1)}}\right\}.
}
\label{eq:reward-age}
\end{equation}
\begin{equation}\medmath{
 \hspace{-2 mm} e_{k,h}^P(s,a)
 \le6\min\left\{1,
 \sqrt{\frac{\Lambda}
 {v(\min\{B,\tau_k^P(s,a)\}H+h+1)}}\right\}.
}
\label{eq:transition-age}
\end{equation}
We define
\begin{equation}
 \begin{gathered}
 A_R:=6\bar R\sqrt{\frac{L}{v}},
 A_P:=6\sqrt{\frac{\Lambda}{v}},
 a_B^j:=\frac{A_j}{\sqrt{BH+1}},j\in\{R,P\}.
 \end{gathered}
 \label{eq:A-and-cap}
\end{equation}
Then, on the same event, for $j\in\{R,P\}$,
\begin{equation}
 e_{k,h}^j(s,a)\le a_B^j+
 A_j(q_{k,h}^j(s,a)+1)^{-1/2}.
 \label{eq:capped-age}
\end{equation}
\end{lemma}
\begin{proof}
Fix $(s,a)$ and $h$, and first consider rewards. Set
$m_*:=\min\{B,\tau_k^R(s,a)\}$ and $q_*:=m_*H+h$.
By definition of $\tau_k^R(s,a)$, every candidate history
with $0\le m\le m_*$ has reward mean $R_k(s,a)$ throughout.
On $\mc G_k$, Lemma~\ref{lem:concentration} therefore
ensures that each corresponding interval contains $R_k(s,a)$.
Therefore, their intersection
$J_{k,h}^{R,(m_*)}(s,a)$ in \eqref{eq:intersection}
contains $R_k(s,a)$ and is nonempty. The selection rule
\eqref{eq:select-reward} thus yields
$\widehat m_{k,h}^R(s,a)\ge m_*$. By
\eqref{eq:reward-estimate}, the estimate lies in the selected
intersection. As these intersections are nested, both
$\widehat R_{k,h}(s,a)$ and $R_k(s,a)$ belong to
$I_{k,h}^{R,(m_*)}(s,a)$. By writing
$N:=N_{k,h}^{(m_*)}(s,a)$, the interval definition in
\eqref{eq:reward-interval}, we have, for $N>0$,
\begin{equation}
 e_{k,h}^R(s,a)
 \le\min\left\{2\bar R,\,
 2\bar R\sqrt{\frac{2L}{N}}\right\}.
 \label{eq:reward-clean-error}
\end{equation}
As the true reward mean and its estimate both lie in
$[-\bar R,\bar R]$, the bound $e_{k,h}^R(s,a)\le2\bar R$
also holds when $N=0$. If $q_*\ge8L/v$, the count condition defining $\mc G_k^N$
in Lemma~\ref{lem:concentration} implies $N\ge vq_*/2$,
as $\mc G_k\subseteq\mc G_k^N$. Hence,
\(
 e_{k,h}^R(s,a)
 \le4\bar R\sqrt{L/vq_*}
 \le6\bar R\sqrt{L/v(q_*+1)},
 \)
where the last step uses $q_*\ge1$. Combining this with
$e_{k,h}^R(s,a)\le2\bar R$ proves \eqref{eq:reward-age}
when $q_*\ge8L/v$. If $q_*<8L/v$, then
$q_*+1\le9L/v$ because $L/v\ge1$. The right-hand side of
\eqref{eq:reward-age} is therefore at least $2\bar R$,
so the deterministic bound also covers this case,
including $q_*=0$. For transitions, set $m_*:=\min\{B,\tau_k^P(s,a)\}$ and
$q_*:=m_*H+h$. The transition coverage in
Lemma~\ref{lem:concentration} and the selection rule
\eqref{eq:select-transition} similarly give
$\widehat m_{k,h}^P(s,a)\ge m_*$. By the nesting in
\eqref{eq:intersection} and the estimator definition in
\eqref{eq:transition-estimate}, both the true and estimated
rows belong to $\mc B_{k,h}^{P,(m_*)}(s,a)$.
For $N:=N_{k,h}^{(m_*)}(s,a)>0$, the triangle inequality
and the radius in \eqref{eq:transition-ball} establish the following inequality:
\begin{equation}
 e_{k,h}^P(s,a)
 \le\min\left\{2,\,
 2\sqrt{\frac{2\Lambda}{N}}\right\},
 \label{eq:transition-clean-error}
\end{equation}
where $e_{k,h}^P(s,a)\le2$ follows because both rows are
probability distributions. If $q_*\ge8L/v$, the count
condition on $\mc G_k^N$ in Lemma~\ref{lem:concentration}
yeilds $N\ge vq_*/2$, and hence
$e_{k,h}^P(s,a)\le4\sqrt{\Lambda/(vq_*)}
\le6\sqrt{\Lambda/[v(q_*+1)]}$.
Together with $e_{k,h}^P(s,a)\le2$, this establishes
\eqref{eq:transition-age} for this case. If $q_*<8L/v$,
then $q_*+1\le9L/v\le9\Lambda/v$, so the right-hand side
of \eqref{eq:transition-age} is at least two. The deterministic
bound therefore also proves the claim in this case,
including when $N=0$. 

Finally, fix $j\in\{R,P\}$. When $\tau_k^j(s,a)<B$,
the corresponding bound in \eqref{eq:reward-age} or
\eqref{eq:transition-age} yields
$e_{k,h}^j(s,a)\le A_j/\sqrt{q_{k,h}^j(s,a)+1}$.
When $\tau_k^j(s,a)\ge B$, it instead gives
$e_{k,h}^j(s,a)\le A_j/\sqrt{BH+h+1}\le a_B^j$,
using \eqref{eq:A-and-cap}. Both cases imply
\eqref{eq:capped-age}.
\end{proof}
On $\mc G_k$, the confidence set indexed by
$\min\{B,\tau_k^j(s,a)\}$ contains both the current component and, by the nesting in \eqref{eq:intersection}, the selected estimate, even when a longer history is selected. Thus, the error bounds require neither change identification nor a minimum change magnitude. After separating $a_B^j$ in \eqref{eq:capped-age}, Lemma~\ref{lem:accumulation} controls
the average of $A_j/(q_{k,h}^j(s,a)+1)^{1/2}$ through the
total number of component changes. Its pathwise bound
holds for all normalized weights satisfying the occupancy
constraint, including those obtained from policies learned
from the same observations.
\subsection{Accumulation across Deployments}
Let $N_{\mathrm{ch}}^R$ and $N_{\mathrm{ch}}^P$ denote the
total numbers of changes to reward entries and transition
rows, respectively, across deployment boundaries. In particular, $N_{\mathrm{ch}}^j:=\sum_{k=0}^{K-2}|\mc C_k^j|$
for $j\in\{R,P\}$, with empty sums equal to zero.
Assumption~\ref{ass:sparse} and Lemma~\ref{lem:rows} give
$N_{\mathrm{ch}}^R\le\epsilon d(K-1)$ and
$N_{\mathrm{ch}}^P\le\vartheta d(K-1)$. Each component remains constant between changes.

\begin{lemma}\textbf{(Weighted Accumulation of Errors)}
\label{lem:accumulation}
Let $T:=KH$ and $A\ge0$. Consider a component process with
ages $q_{k,h}(s,a)$ as defined above and $N_{\mathrm{ch}}$
component changes. Suppose the nonnegative errors satisfy
\(
 f_{k,h}(s,a)\le A(q_{k,h}(s,a)+1)^{-1/2},
 \)
and the weights satisfy
\begin{equation}
 \sum_{s,a}w_{k,h}(s,a)=1,
 \quad 0\le w_{k,h}(s,a)\le\frac{\kappa}{d}.
 \label{eq:weights}
\end{equation}
Then, we have
\begin{equation}
 \begin{aligned}
 &\frac1T\sum_{k=0}^{K-1}\sum_{h=0}^{H-1}
 \sum_{s,a}w_{k,h}(s,a)f_{k,h}(s,a) \le\frac{2A}{\sqrt T}
 +2A\sqrt{\frac{\kappa N_{\mathrm{ch}}}{dT}}.
 \end{aligned}
 \label{eq:accumulation}
\end{equation}
\end{lemma}
\begin{proof}
Reindex time by $t=kH+h$. Let $b_t:=A/\sqrt{t+1}$ and
\(z_t(s,a):=\max\{f_t(s,a)-b_t,\,0\}\). Since $f_t(s,a)\le b_t+z_t(s,a)$
and the weights sum to one, the contribution of $b_t$ to the
weighted average is at most $T^{-1}\sum_t b_t\le 2A/\sqrt T$.
Before the first change of $(s,a)$, we have $q_t(s,a)=t$, so
$z_t(s,a)=0$. In each interval beginning at a component change,
the age starts at zero and increases by one per time step.
Hence $z_t(s,a)>u$ implies $q_t(s,a)+1<A^2/u^2$, which can
occur at most $A^2/u^2$ times per interval. Summing over the
$N_{\mathrm{ch}}$ such intervals and using the weight constraints
gives $T^{-1}\sum_{t,s,a}w_t(s,a)\mathbf 1\{z_t(s,a)>u\}
\le \min\{1,\kappa N_{\mathrm{ch}}A^2/(dTu^2)\}$.
Integrating over $u>0$ bounds the weighted average of $z_t$
by $2A\sqrt{\kappa N_{\mathrm{ch}}/(dT)}$. Adding the two
contributions proves the claim.
\end{proof}
The proof uses only the error bound on $f_{k,h}(s,a)$ and the
weight constraints in \eqref{eq:weights}. Its conclusion therefore holds for every realization satisfying
these conditions.
\subsection{From Model Error to Policy Loss}
The following lemma bounds the loss in the true MDP of a
policy optimal for the estimated model by a weighted sum
of reward and transition errors. Under
Assumption~\ref{ass:occupancy}, the resulting occupancy
weights satisfy \eqref{eq:weights}, allowing us to apply
Lemma~\ref{lem:accumulation}.
\begin{lemma}\textbf{(Policy comparison for an estimated model)}
\label{lem:comparison}
Fix a deployment $k$. Let $(\widehat{\mc P},\widehat R)$
be an estimated model on the same state and action spaces,
satisfying $\widehat{\mc P}(\cdot\mid s,a)\in\Delta(\mc S)$
and $|\widehat R(s,a)|\le\bar R$ for every $(s,a)$.
Let $\widehat\pi$ be an optimal policy for this model.
Define the reward and transition errors by
$e_R(s,a):=|R_k(s,a)-\widehat R(s,a)|$ and
$e_P(s,a):=\norm*{\mc P_k(\cdot\mid s,a)
-\widehat{\mc P}(\cdot\mid s,a)}_1$, respectively. Then
\begin{equation}
 \begin{aligned}
 \ell_k^\rho(\widehat\pi)
 &\le\frac{2}{1-\gamma}\sum_{s,a}w(s,a)\cdot\left[e_R(s,a)
 +\frac{\gamma\bar R}{1-\gamma}e_P(s,a)\right],
 \end{aligned}
 \label{eq:comparison}
\end{equation}
where
\(
 w(s,a):=\frac12\left[
 d_{\rho,k}^{\pi_k^*}(s,a)
 +d_{\rho,k}^{\widehat\pi}(s,a)\right].
 \)
Under Assumption~\ref{ass:occupancy}, these weights satisfy
\eqref{eq:weights}.
\end{lemma}
\begin{proof}
For any stationary policy $\pi$, let $\widehat V^\pi$ be its
value in the fitted MDP and set
$\widehat J^\rho(\pi):=\rho^\top\widehat V^\pi$.
Subtracting the Bellman equations and expanding the true-model
resolvent gives
$J_k^\rho(\pi)-\widehat J^\rho(\pi)
=(1-\gamma)^{-1}\sum_{s,a}d_{\rho,k}^\pi(s,a)
\big[(R_k-\widehat R)(s,a)
+\gamma\langle(\mc P_k-\widehat{\mc P})(\cdot\mid s,a),
\widehat V^\pi\rangle\big]$.
Using $\norm{\widehat V^\pi}_\infty\le\bar R/(1-\gamma)$ yields the following bound:
\begin{equation}\medmath{
|J_k^\rho(\pi)-\widehat J^\rho(\pi)|
\le \frac{1}{1-\gamma}\sum_{s,a}d_{\rho,k}^\pi(s,a)
\left[e_R(s,a)+\frac{\gamma\bar R}{1-\gamma}e_P(s,a)\right].
}
\label{eq:fixed-policy-comparison}
\end{equation}
Since $\widehat J^\rho(\pi_k^*)\le\widehat J^\rho(\widehat\pi)$,
the policy loss is bounded by the sum of the absolute return
errors for $\pi_k^*$ and $\widehat\pi$. Applying
\eqref{eq:fixed-policy-comparison} to both policies gives
\eqref{eq:comparison} with their average occupancy
\(w(s,a)\). Normalization and
Assumption~\ref{ass:occupancy} give \eqref{eq:weights}.
\end{proof}
\begin{proof}[\(\bullet\)~\textbf{Proof of Theorem~\ref{thm:main}}]
Set $T:=KH$, and take the weights $w_{k,h}(s,a)$ from Lemma~\ref{lem:comparison}, satisfying \eqref{eq:weights}. For $j\in\{R,P\}$, define $f_{k,h}^j(s,a):=(e_{k,h}^j(s,a)-a_B^j)_+\mathbf1_{\mc G_k}$. Leveraging Lemma~\ref{lem:adaptive}, we get $0\le f_{k,h}^j(s,a)\le A_j/\sqrt{q_{k,h}^j(s,a)+1}$ for every realization. Since $e_{k,h}^j(s,a)\mathbf1_{\mc G_k}\le a_B^j+f_{k,h}^j(s,a)$, weight normalization and Lemma~\ref{lem:accumulation} yield $T^{-1}\sum_{k,h}\sum_{s,a}w_{k,h}(s,a)e_{k,h}^j(s,a)\mathbf1_{\mc G_k}\le D_j$, where $D_j:=a_B^j+2A_j/\sqrt T+2A_j\sqrt{\kappa N_{\mathrm{ch}}^j/(dT)}$. Apply \eqref{eq:comparison} on $\mc G_k$ and the loss cap $2\bar R/(1-\gamma)$ on $\mc G_k^c$. Averaging and taking expectations, using $\mathbb P(\mc G_k^c)\le\delta$ from Lemma~\ref{lem:concentration}, gives $\mathbb E[\mc L_{K,H}^\rho]\le 2D_R/(1-\gamma)+2\gamma\bar R D_P/(1-\gamma)^2+2\bar R\delta/(1-\gamma)$. Assumption~\ref{ass:sparse} and Lemma~\ref{lem:rows} imply $N_{\mathrm{ch}}^R\le\epsilon d(K-1)$ and $N_{\mathrm{ch}}^P\le\vartheta d(K-1)$. Substituting these bounds and \eqref{eq:A-and-cap} into $D_j$, then applying Cauchy--Schwarz, yields $D_R\le C\bar R\sqrt{L(\bullet+\kappa\epsilon)/(vH)}$ and $D_P\le C\sqrt{\Lambda(\bullet+\kappa\vartheta)/(vH)}$, where $\bullet:=B^{-1}+K^{-1}$ and $C$ is an absolute constant. Inserting these estimates proves \eqref{eq:main-bound}.
\end{proof}
\begin{proof}[\(\bullet\)~\textbf{Proof of Theorem~\ref{cor:persistent}}]
The choice $B=\ceil{\epsilon^{-1}}$ gives
$B^{-1}\le\epsilon$ and $B+1\le3/\epsilon$.
With $\delta=\epsilon/(H+1)$, \eqref{eq:logs} therefore
implies $L\le L_\epsilon$ and $\Lambda\le\Lambda_\epsilon$.
As $\kappa\ge1$ and $\vartheta\ge\epsilon/2$, we also have
$B^{-1}+\kappa\epsilon\le2\kappa\epsilon$ and
$B^{-1}+\kappa\vartheta\le3\kappa\vartheta$.
Substituting these estimates into the separate reward and
transition bounds established in the proof of
Theorem~\ref{thm:main} gives \eqref{eq:persistent-finite},
after absorbing the additive failure term into the reward
term using
$\delta=\epsilon/(H+1)
\le\sqrt{\kappa\epsilon L_\epsilon/(vH)}$.
Both parameter choices are independent of $K$, so the same
learner retains $\mathcal O(\epsilon^{-1})$ deployment
summaries for every horizon. Taking $\limsup_{K\to\infty}$
in \eqref{eq:persistent-finite} proves
\eqref{eq:persistent-longrun}.
\end{proof}
\section{Conclusion}
We studied RL in discounted MDPs with sparse changes in reward and transition entries between deployments. We proposed an approach for the learner that uses confidence sets to select how much past history to pool separately for each reward mean and transition row. Under independent sampling and $0<|\mc S|\epsilon\le 2$, we demonstrated that retaining $\mathcal O(\epsilon^{-1})$ deployment summaries suffices to achieve a long-run expected average policy loss of $\widetilde{\mathcal O}(\sqrt{\epsilon/H})$.
\bibliographystyle{unsrt} 
\bibliography{refs}
\end{document}